\documentclass[letterpaper, 10 pt, conference]{ieeeconf}

\IEEEoverridecommandlockouts

\usepackage{graphicx}
\usepackage{amsmath,amssymb}
\usepackage{subfig}
\usepackage{bm}
\usepackage{epstopdf}
\usepackage[table]{xcolor}
\usepackage{flushend}

\definecolor{bedColor}{rgb}{0.0,0.501960784314,0.501960784314}
\definecolor{booksColor}{rgb}{0.980392156863,0.196078431373,0.196078431373}
\definecolor{ceilColor}{rgb}{0.4,0.0,0.8}
\definecolor{chairColor}{rgb}{0.196078431373,0.196078431373,0.980392156863}
\definecolor{floorColor}{rgb}{0.9,0.9,0.93}
\definecolor{furnColor}{rgb}{1.0,0.270588235294,0.078431372549}
\definecolor{objsColor}{rgb}{1.0,0.078431372549,0.498039215686}
\definecolor{paintColor}{rgb}{0.196078431373,0.196078431373,0.588235294118}
\definecolor{sofaColor}{rgb}{0.870588235294,0.705882352941,0.549019607843}
\definecolor{tableColor}{rgb}{0.196078431373,0.980392156863,0.196078431373}
\definecolor{tvColor}{rgb}{1.0,0.843137254902,0.0}
\definecolor{wallColor}{rgb}{0.588235294118,0.588235294118,0.588235294118}
\definecolor{windColor}{rgb}{0.0,1.0,1.0}

\title{\LARGE \bf
SemanticFusion: Dense 3D Semantic Mapping with Convolutional Neural Networks
}

\author{John McCormac, Ankur Handa, Andrew Davison, and Stefan Leutenegger\\
Dyson Robotics Lab, Imperial College London
}

\begin{document}

\maketitle
\thispagestyle{empty}
\pagestyle{empty}

\begin{abstract}

Ever more robust, accurate and detailed mapping using visual sensing has proven to be an enabling factor for mobile robots across a wide variety of applications. For the next level of robot intelligence and intuitive user interaction, maps need extend beyond geometry and appearence --- they need to contain semantics. We address this challenge by combining Convolutional Neural Networks (CNNs) and a state of the art dense Simultaneous Localisation and Mapping (SLAM) system, ElasticFusion, which provides long-term dense correspondence between frames of indoor {RGB-D} video even during loopy scanning trajectories.  These correspondences allow the CNN's semantic predictions from multiple view points to be probabilistically fused into a map.  This not only produces a useful semantic 3D map, but we also show on the NYUv2 dataset that fusing multiple predictions leads to an improvement even in the 2D semantic labelling over baseline single frame predictions. We also show that for a smaller reconstruction dataset with larger variation in prediction viewpoint, the improvement over single frame segmentation increases.  Our system is efficient enough to allow real-time interactive use at frame-rates of $\approx$25Hz.

\end{abstract}

\section{INTRODUCTION}

The inclusion of rich semantic information within a dense map enables a much greater range of functionality than geometry alone. For instance, in domestic robotics, a simple fetching task requires knowledge of both what something is, as well as where it is located. As a specific example, thanks to sharing of the same spatial and semantic understanding between user and robot, we may issue commands such as 'fetch the coffee mug from the nearest table on your right'. Similarly, the ability to query semantic information within a map is useful for humans directly, providing a database for answering spoken queries about the semantics of a previously made map; `How many chairs do we have in the conference room? What is the distance between the lectern and its nearest chair?' In this work, we combine the geometric information from a state-of-the-art SLAM system ElasticFusion \cite{Whelan:etal:RSS2015} with recent advances in semantic segmentation using Convolutional Neural Networks (CNNs).

\begin{figure}
\centering
\includegraphics[width=8.2cm]{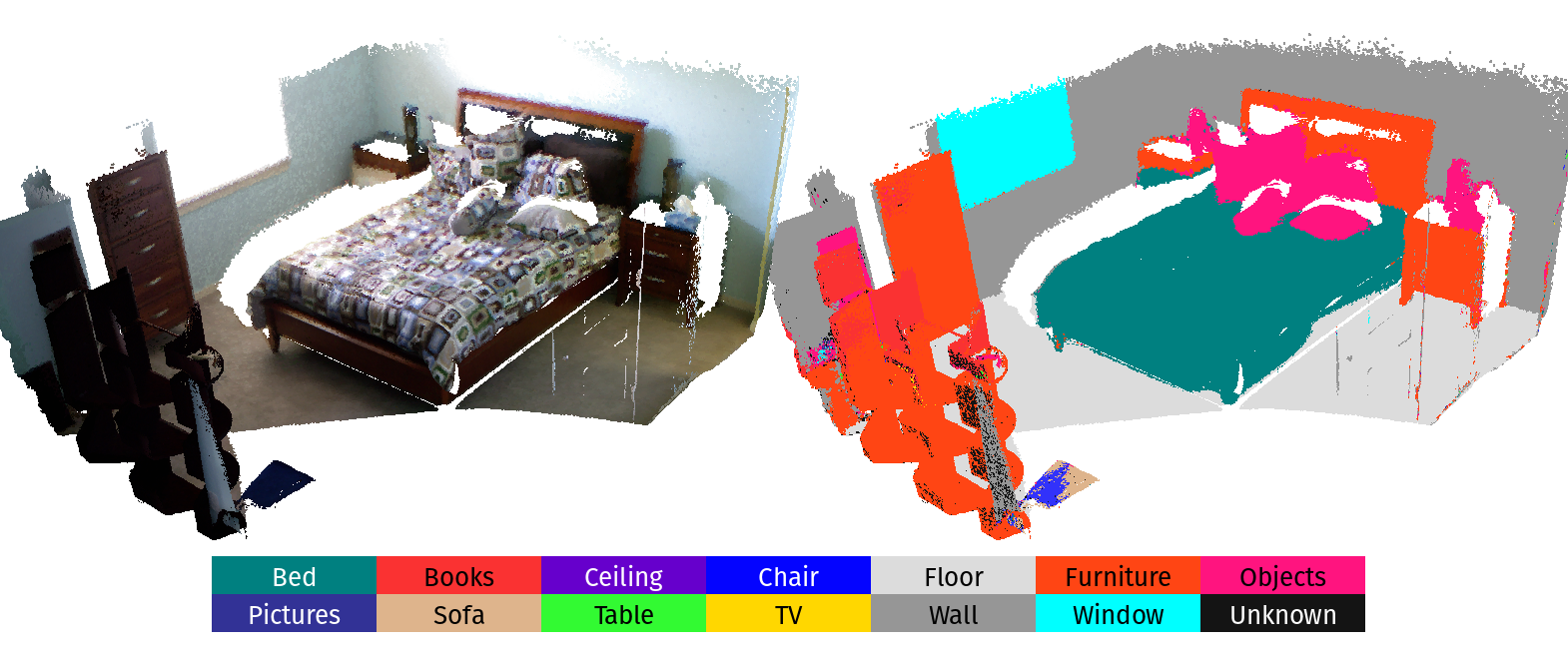}\label{fig:combined_annotated_reconstruction}
\caption{\textbf{The output of our system}: On the left, a dense surfel based reconstruction from a video sequence in the NYUv2 test set. On the right the same map, semantically annotated with the classes given in the legend below.}
\vspace{2mm}\hrule 
\label{fig:system_output}
\end{figure}

Our approach is to use the SLAM system to provide correspondences from the 2D frame into a globally consistent 3D map.  This allows the CNN's semantic predictions from multiple viewpoints to be probabilistically fused into a dense semantically annotated map, as shown in Figure~\ref{fig:system_output}. ElasticFusion is particularly suitable for fusing semantic labels because its surfel-based surface representation is automatically deformed to remain consistent after the small and large loop closures which would frequently occur during typical interactive use by an agent (whether human or robot). As the surface representation is deformed and corrected, individual surfels remain persistently associated with real-world entities and this enables long-term fusion of per-frame semantic predictions over wide changes in viewpoint.

The geometry of the map itself also provides useful information which can be used to efficiently regularise the final predictions.  Our pipeline is designed to work online, and although we have not focused on performance, the efficiency of each component leads to a real-time capable ($\approx~25$Hz) interactive system.  The resulting map could also be used as a basis for more expensive offline processing to further improve both the geometry and the semantics; however that has not been explored in the current work.

We evaluate the accuracy of our system on the NYUv2 dataset, and show that by using information from the unlabelled raw video footage we can improve upon baseline approaches performing segmentation using only a single frame. This suggests the inclusion of SLAM not only provides an immediately useful semantic 3D map, but it suggests that many state-of-the art 2D single frame semantic segmentation approaches may be boosted in performance when linked with SLAM. 

The NYUv2 dataset was not taken with full room reconstruction in mind, and often does not provide significant variations in viewpoints for a given scene. To explore the benefits of SemanticFusion within a more thorough reconstruction, we developed a small dataset of a reconstructed office room, annotated with the NYUv2 semantic classes.  Within this dataset we witness a more significant improvement in segmentation accuracy over single frame 2D segmentation.  This indicates that the system is particularly well suited to longer duration scans with wide viewpoint variation aiding to disambiguate the single-view 2D semantics.

\section{RELATED WORK}

The works most closely related are St{\"u}ckler \textit{et al.}~\cite{Stuckler:etal:JRTIP2015} and Hermans \textit{et al.}~\cite{Hermans:etal:ICRA2014}; both aim towards a dense, semantically annotated 3D map of indoor scenes.  They both obtain per-pixel label predictions for incoming frames using Random Decision Forests, whereas ours exploits recent advances in Convolutional Neural Networks that provide state-of-the-art accuracy, with a real-time capable run-time performance.  They both fuse predictions from different viewpoints in a classic Bayesian framework.  St{\"u}ckler \textit{et al.}~\cite{Stuckler:etal:JRTIP2015} used a Multi-Resolution Surfel Map-based {SLAM} system capable of operating at 12.8Hz, however unlike our system they do not maintain a single global semantic map as local key frames store aggregated semantic information and these are subject to graph optimisation in each frame. Hermans \textit{et al.}~\cite{Hermans:etal:ICRA2014} did not use the capability of a full SLAM system with explicit loop closure: they registered the predictions in the reference frames using only camera tracking. Their run-time performance was 4.6Hz, which would prohibit processing a live video feed, whereas our system is capable of operating online and interactively. As here, they regularised their predictions using Kr\"ahenb\"uhl and Koltun's~\cite{Krahenbuhl:Koltun:NIPS2011} fully-connected CRF inference scheme to obtain a final semantic map.

Previous work by Salas-Moreno \textit{et al.} aimed to create a fully capable SLAM system, SLAM\verb!++!~\cite{Salas-Moreno:etal:CVPR2013}, which maps indoor scenes at the level of semantically defined objects. However, their method is limited to mapping objects that are present in a pre-defined database. It also does not provide the dense labelling of entire scenes that we aim for in this work, which also includes walls, floors, doors, and windows which are equally important to describe the extent of the room. Additionally, the features they use to match template models are hand-crafted unlike our CNN features that are learned in an end-to-end fashion with large training datasets.

The majority of other approaches to indoor semantic labelling either focuses on offline batch mapping methods~\cite{Valentin:etal:CVPR2014,Koppula:etal:NIPS2011} or on single-frame 2D segmentations which do not aim to produce a semantically annotated 3D map~\cite{Everingham:etal:IJCV2010,Silberman:etal:ECCV2012,Lin:etal:ECCV2014,Song:etal:CVPR2015}.  Valentin \textit{et al.}~\cite{Valentin:etal:CVPR2014} used a CRF and a per-pixel labelling from a variant of TextonBoost to reconstruct semantic maps of both indoor and outdoor scenes. This produces a globally consistent 3D map, however inference is performed on the whole mesh once instead of incrementally fusing the predictions online. Koppula \textit{et al.} ~\cite{Koppula:etal:NIPS2011} also tackle the problem on a completed 3D map, forming segments of the map into nodes of a graphical model and using hand-crafted geometric and visual features as edge potentials to infer the final semantic labelling.

Our semantic mapping pipeline is inspired by the recent success of Convolution Neural Networks in semantic labelling and segmentation tasks~\cite{Krizhevsky:etal:NIPS2012,Long:etal:CVPR2015,Noh:etal:ARXIV2015}.  CNNs have proven capable of both state-of-the-art accuracy and efficient test-time performance. They have have exhibited these capabilities on numerous datasets and a variety of data modalities, in particular RGB~\cite{Noh:etal:ARXIV2015,Long:etal:CVPR2015}, Depth~\cite{Couprie:etal:ICLR2013,Handa:etal:ARXIV2015} and Normals~\cite{Eigen:etal:ICCV2015,Gupta:etal:ECCV2014,Gupta:etal:CVPR2015,Gupta:etal:CVPR2015B}. In this work we build on the CNN model proposed by Noh \textit{et. al.}~\cite{Noh:etal:ARXIV2015}, but modify it to take advantage of the directly available depth data in a manner that does not require significant additional pre-processing.

\section{METHOD}

\begin{figure}[t]
\centering
\includegraphics[width=8.2cm]{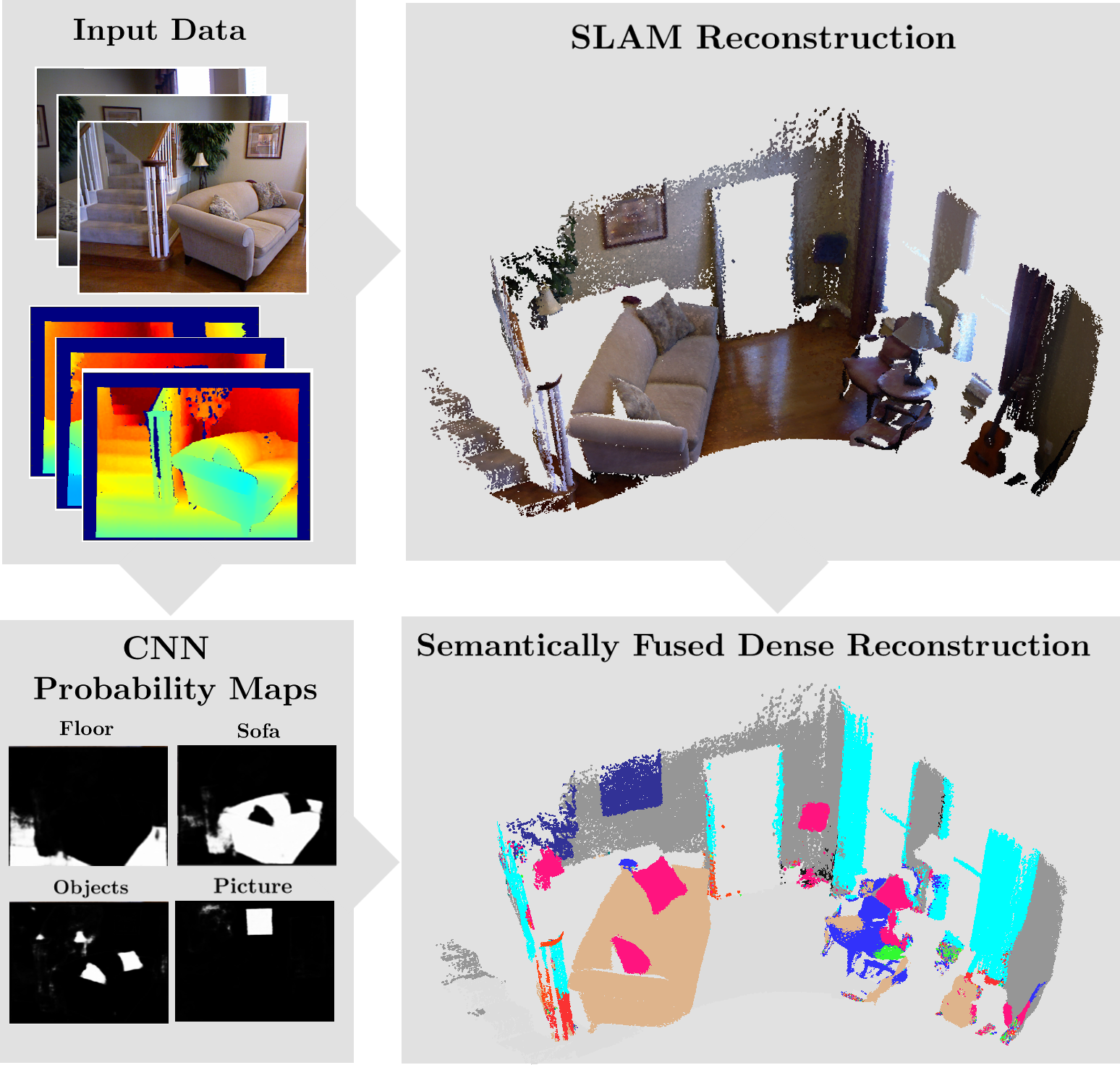}
\caption{\textbf{An overview of our pipeline}: Input images are used to produce a SLAM map, and a set of probability prediction maps (here only four are shown).  These maps are fused into the final dense semantic map via Bayesian updates.}
\vspace{2mm}\hrule 
\label{fig:full_pipeline}
\end{figure}

Our SemanticFusion pipeline is composed of three separate units; a real-time SLAM system ElasticFusion, a Convolutional Neural Network, and a Bayesian update scheme, as illustrated in Figure~\ref{fig:full_pipeline}.  The role of the SLAM system is to provide correspondences between frames, and a globally consistent map of fused surfels.  Separately, the CNN receives a 2D image (for our architecture this is RGBD, for Eigen \textit{et al.}~\cite{Eigen:etal:ICCV2015} it also includes estimated normals), and returns a set of per pixel class probabilities.  Finally, a Bayesian update scheme keeps track of the class probability distribution for each surfel, and uses the correspondences provided by the SLAM system to update those probabilities based on the CNN's predictions.  Finally, we also experiment with a CRF regularisation scheme to use the geometry of the map itself to improve the semantic predictions \cite{Hermans:etal:ICRA2014,Krahenbuhl:Koltun:NIPS2011}.  The following section outlines each of these components in more detail.

\subsection{SLAM Mapping}

We choose ElasticFusion as our SLAM system.\footnote{Available on \texttt{https://github.com/mp3guy/ElasticFusion}} 
For each arriving frame, $k$, ElasticFusion tracks the camera pose via a combined ICP and RGB alignment, to yield a new pose $\bm T_{WC}$, where $W$ denotes the World frame and $C$ the camera frame. 
New surfels are added into our map using this camera pose, and existing surfel information is combined with new evidence to refine their positions, normals, and colour information. Additional checks for a loop closure event run in parallel and the map is optimised immediately upon a loop closure detection. 

The deformation graph and surfel based representation of ElasticFusion lend themselves naturally to the task at hand, allow probability distributions to be `carried along' with the surfels during loop closure, and also fusing new depth readings to update the surfel's depth and normal information, without destroying the surfel, or its underlying probability distribution.  It operates at real-time frame-rates at VGA resolution and so can be used both interactively by a human or in robotic applications. We used the default parameters in the public implementation, except for the depth cutoff, which we extend from 3m to 8m to allow reconstruction to occur on sequences with geometry outside of the 3m range.

\subsection{CNN Architecture}
\label{sec:cnn_method}

Our CNN is implemented in \textit{caffe} \cite{Jia:etal:ARXIV2014} and adopts the Deconvolutional Semantic Segmentation network architecture proposed by Noh \textit{et. al.}~\cite{Noh:etal:ARXIV2015}.  Their architecture is itself based on the VGG 16-layer network~\cite{Simonyan:Zisserman:ICLR2015}, but with the addition of max unpooling and deconvolutional layers which are trained to output a dense pixel-wise semantic probability map.  This CNN was trained for RGB input, and in the following sections when using a network with this setup we describe it RGB-CNN.

Given the availability of depth data, we modified the original network architecture to accept depth information as a fourth channel. Unfortunately, the depth modality lacks the large scale training datasets of its RGB counterpart.  The NYUv2 dataset only consists of 795 labelled training images.  To effectively use depth, we initialized the depth filters with the average intensity of the other three inputs, which had already been trained on a large dataset, and converted it from the 0--255 colour range to the 0--8m depth range by increasing the weights by a factor of $\approx32\times$.

We rescale incoming images to the native 224$\times$224 resolution for our CNNs, using bilinear interpolation for RGB, and nearest neighbour for depth. In our experiments with Eigen \textit{et. al.'s} implementation we rescale the inputs in the same manner to 320$\times$240 resolution.  We upsample the network output probabilites to full 640$\times$480 image resolution using nearest neighbour when applying the update to surfels, described in the section below.

\subsection{Incremental Semantic Label Fusion}

In addition to normal and location information, each surfel (index $s$) in our map, $\mathcal M$, stores a discrete probability distribution, $P(L_s = l_i)$ over the set of class labels, $l_i \in \mathcal{L}$. 
Each newly generated surfel is initialised with a uniform distribution over the semantic classes, as we begin with no \emph{a priori} evidence as to its latent classification.  

After a prespecified number of frames, we perform a forward pass of the CNN with the image $\bm I_k$ coming directly from the camera. Depending on the CNN architecture, this image can include any combination of RGB, depth, or normals.
Given the data $\bm I_k$ of the $k^\mathrm{th}$ image, the output of the CNN is interpreted in a simplified manner as a per-pixel independent probability distribution over the class labels $P(O_{\bm u} = l_i | \bm I_k)$, with $\bm u$ denoting pixel coordinates.

Using the tracked camera pose $\bm T_{WC}$, we associate every surfel at a given 3D location $_W\bm x(s)$ in the map, with pixel coordinates $\bm u$ via the camera projection $\bm u(s,k) = \bm \pi (\bm T_{CW}(k) _W\bm x(s))$, employing the homogeneous transformation matrix $\bm T_{CW}(k) = \bm T_{WC}^{-1}(k)$ and using homogeneous 3D coordinates.
This enables us to update all the surfels in the visible set $\mathcal V_k \subseteq \mathcal M$ with the corresponding probability distribution by means of a recursive Bayesian update

\begin{equation}
\begin{split}
	P(l_i | \bm I_{1,\ldots,k}) = \frac{1}{Z} P(l_i | \bm I_{1,\ldots,k-1}) P(O_{\bm u(s,k)} = l_i | \bm I_k),
\end{split}
\end{equation}

which is applied to all label probabilities per surfel, finally normalising with constant $Z$ to yield a proper distribution.

It is the SLAM correspondences that allow us to accurately associate label hypotheses from multiple images and combine evidence in a Bayesian way. 
The following section discusses how the na\"{i}ve independence approximation employed so far can be mitigated, allowing semantic information to be propagated spatially when semantics are fused from different viewpoints.

\subsection{Map Regularisation}

We explore the benefits of using map geometry to regularise predictions by applying a fully-connected CRF with Gaussian edge potentials to surfels in the 3D world frame, as in the work of Hermans \textit{et al.}~\cite{Hermans:etal:ICRA2014,Krahenbuhl:Koltun:NIPS2011}.  
We do not use the CRF to arrive at a final prediction for each surfel, but instead use it incrementally to update the probability distributions. In our work, we treat each surfel as a node in the graph. The algorithm uses the mean-field approximation and a message passing scheme to efficiently infer the latent variables that approximately minimise the Gibbs energy $E$ of a labelling, $\mathbf{x}$, in a fully-connected graph, where $x_s \in \{l_i\}$ denotes a given labelling for the surfel with index $s$.

The energy $E(\mathbf{x})$ consists of two parts, the unary data term $\psi_u(x_s) $ is a function of a given label, and is parameterised by the internal probability distribution of the surfel from fusing multiple CNN predictions as described above. The pairwise smoothness term, $\psi_p(x_s,x_{s'})$ is a function of the labelling of two connected surfels in the graph, and is parameterised by the geometry of the map:

\begin{equation}
E(\mathbf{x}) = \sum_s \psi_u(x_s) + \sum_{s<s'} \psi_p(x_s,x_{s'}).
\end{equation}

For the data term we simply use the negative logarithm of the chosen labelling's probability for a given surfel,

\begin{equation}
\psi_u(x_s)= -\text{log}(P(L_s = x_s | \bm I_{1,\ldots,k})).
\end{equation}

In the scheme proposed by Kr\"ahenb\"uhl and Koltun~\cite{Krahenbuhl:Koltun:NIPS2011} the smoothness term is constrained to be a linear combination of $K$ Gaussian edge potential kernels, where $\mathbf{f}_s$ denotes some feature vector for surfel, $s$, and in our case $\mu(x_s,x_{s'})$ is given by the Potts model, $\mu(x_s,x_{s'}) = [x_s \neq x_{s'}]$:

\begin{equation}
\psi_p(x_s,x_{s'}) = \mu(x_s,x_{s'})\left(\sum_{m=1}^K w^{(m)}k^{(m)}(\mathbf{f}_s,\mathbf{f}_{s'})\right).
\end{equation}

Following previous work~\cite{Hermans:etal:ICRA2014} we use two pairwise potentials; a bilateral appearance potential seeking to closely tie together surfels with both a similar position and appearance, and a spatial smoothing potential which enforces smooth predictions in areas with similar surface normals:

\begin{equation}
k^1(\mathbf{f}_s,\mathbf{f}_{s'}) = \text{exp}\left(-\frac{|\mathbf{p}_s - \mathbf{p}_{s'}|^2}{2\theta_\alpha^2}-\frac{|\mathbf{c}_s - \mathbf{c}_{s'}|^2}{2\theta_\beta^2}\right),
\end{equation}

\begin{equation}
k^2(\mathbf{f}_s,\mathbf{f}_{s'}) = \text{exp}\left(-\frac{|\mathbf{p}_s - \mathbf{p}_{s'}|^2}{2\theta_\alpha^2}-\frac{|\mathbf{n}_s - \mathbf{n}_{s'}|^2}{2\theta_\gamma^2}\right).
\end{equation}

We chose unit standard deviations of $\theta_\alpha = 0.05$m in the spatial domain,  $\theta_\beta = 20$ in the RGB colour domain, and $\theta_\gamma = 0.1$ radians in the angular domain. We did not tune these parameters for any particular dataset.  We also maintained $w^1$ of 10 and $w^2$ of 3 for all experiments.  These were the default settings in Kr\"ahenb\"uhl and Koltun's public implementation\footnote{Available from: \texttt{http://www.philkr.net/home/densecrf}}~\cite{Krahenbuhl:Koltun:NIPS2011} .

\section{EXPERIMENTS}

\subsection{Network Training}

We initialise our CNNs with weights from Noh \textit{et. al.}~\cite{Noh:etal:ARXIV2015} trained for segmentation on the PASCAL VOC 2012 segmentation dataset~\cite{Everingham:etal:IJCV2010}. For depth input we initialise the fourth channel as described in Section~\ref{sec:cnn_method}, above. We finetuned this network on the training set of the NYUv2 dataset for the 13 semantic classes defined by Couprie \textit{et al.}~\cite{Couprie:etal:ICLR2013}.

For optimisation we used standard stochastic gradient descent, with a learning rate of 0.01, momentum of 0.9, and weight decay of $5\times10^{-4}$. After 10k iterations we reduced the learning rate to $1\times10^{-3}$.  We use a mini-batch size of 64, and trained the networks for a total of 20k iterations over the course of 2 days on an Nvidia GTX Titan X.

\subsection{Reconstruction Dataset}

\begin{figure}
\centering
\includegraphics[width=8.5cm]{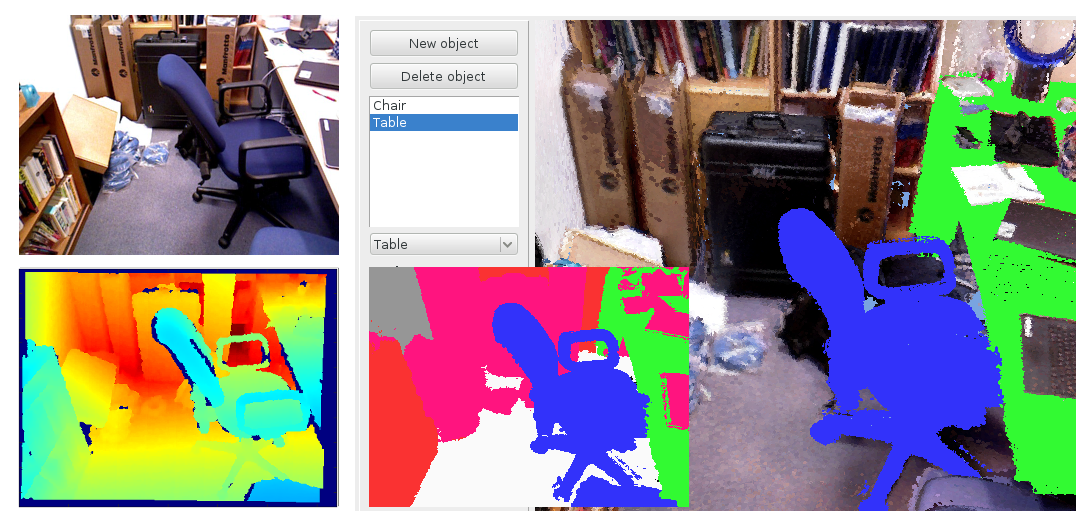}
\caption{\textbf{Our office reconstruction dataset}: On the left are the captured RGB and Depth images.  On the right, is our 3D reconstruction and annotation.  Inset into that is the final ground truth rendered labelling we use for testing.}
\label{fig:office_reconstruction_annotation}
\vspace{2mm}\hrule 
\end{figure}

We produced a small experimental RGB-D reconstruction dataset, which aimed for a relatively complete reconstruction of an office room. The trajectory used is notably more loopy, both locally and globally, than the NYUv2 dataset which typically consists of a single back and forth sweep. We believe the trajectory in our dataset is more representative of the scanning motion an active agent may perform when inspecting a scene. 

We also took a different approach to manual annotation of this data, by using a 3D tool we developed to annotate the surfels of the final 3D reconstruction with the 13 NYUv2 semantic classes under consideration (only 9 were present). We then automatically generated 2D labellings for any frame in the input sequence via projection.  The tool, and the resulting annotations are depicted in Figure~\ref{fig:office_reconstruction_annotation}. Every 100\textsuperscript{th} frame of the sequence was used as a test sample to validate our predictions against the annotated ground truth, resulting in 49 test frames.

\subsection{CNN and CRF Update Frequency Experiments}

\begin{figure}
\centering
\includegraphics[width=8.2cm]{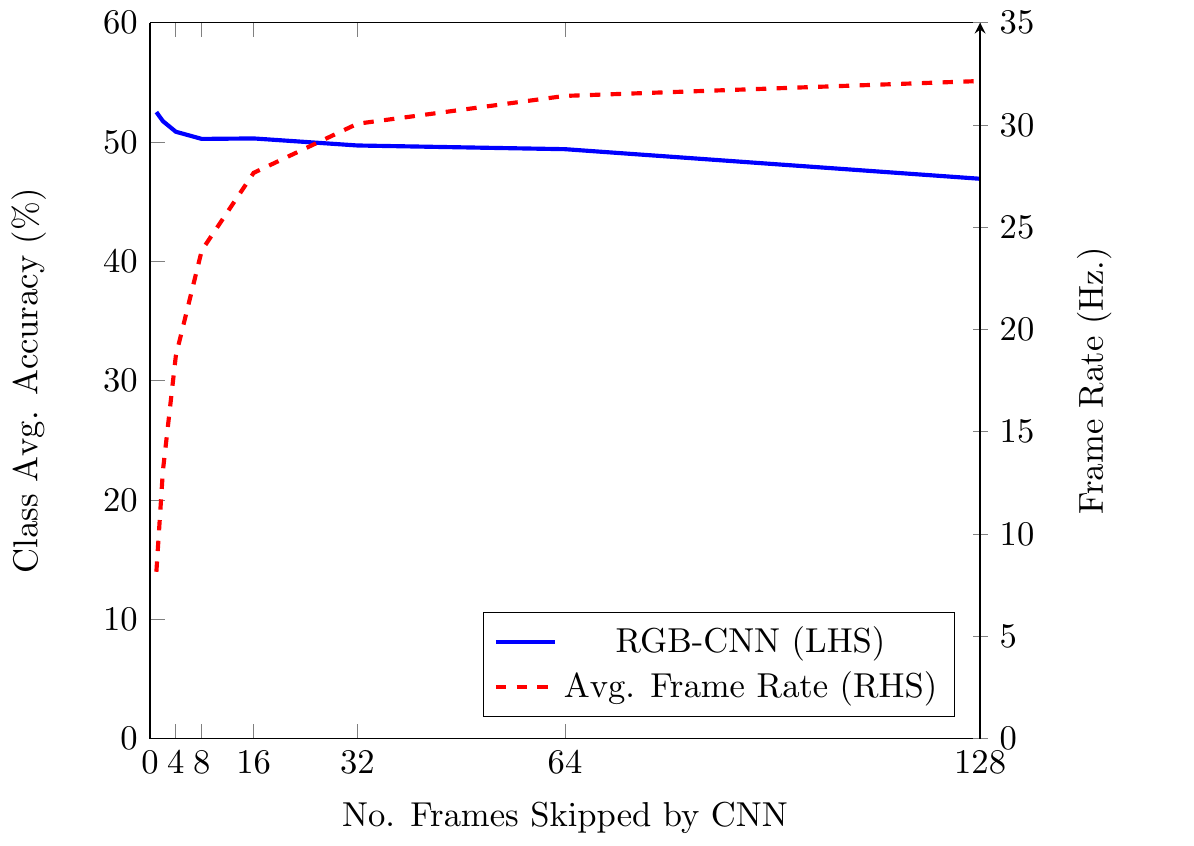}
\caption{The class average accuracy of our RGB-CNN on the office reconstruction dataset against the number of frames skipped between fusing semantic predictions. We perform this evaluation without CRF smoothing. The right hand axis shows the estimated run-time performance in terms of FPS.}
\vspace{2mm}\hrule 
\label{fig:cnn_frameskip_vs_accuracy}
\end{figure}

We used the dataset to evaluate the accuracy of our system when only performing a CNN prediction on a subset of the incoming video frames.  We used the RGB-CNN described above, and evaluated the accuracy of our system when performing a prediction on every $2^n$ frames, where $n \in \{0..7\}$.  We calculate the average frame-rate based upon the run-time analysis discussed in Section~\ref{sec:runtime_analysis}. As shown in Figure~\ref{fig:cnn_frameskip_vs_accuracy}, the accuracy is highest (52.5\%) when every frame is processed by the network, however this leads to a significant drop in frame-rate to 8.2Hz.  Processing every 10\textsuperscript{th} frame results in a slightly reduced accuracy (49-51\%), but over three times the frame-rate of 25.3Hz.  This is the approach taken in all of our subsequent evaluations.

\begin{figure}
\centering
\includegraphics[width=8.2cm]{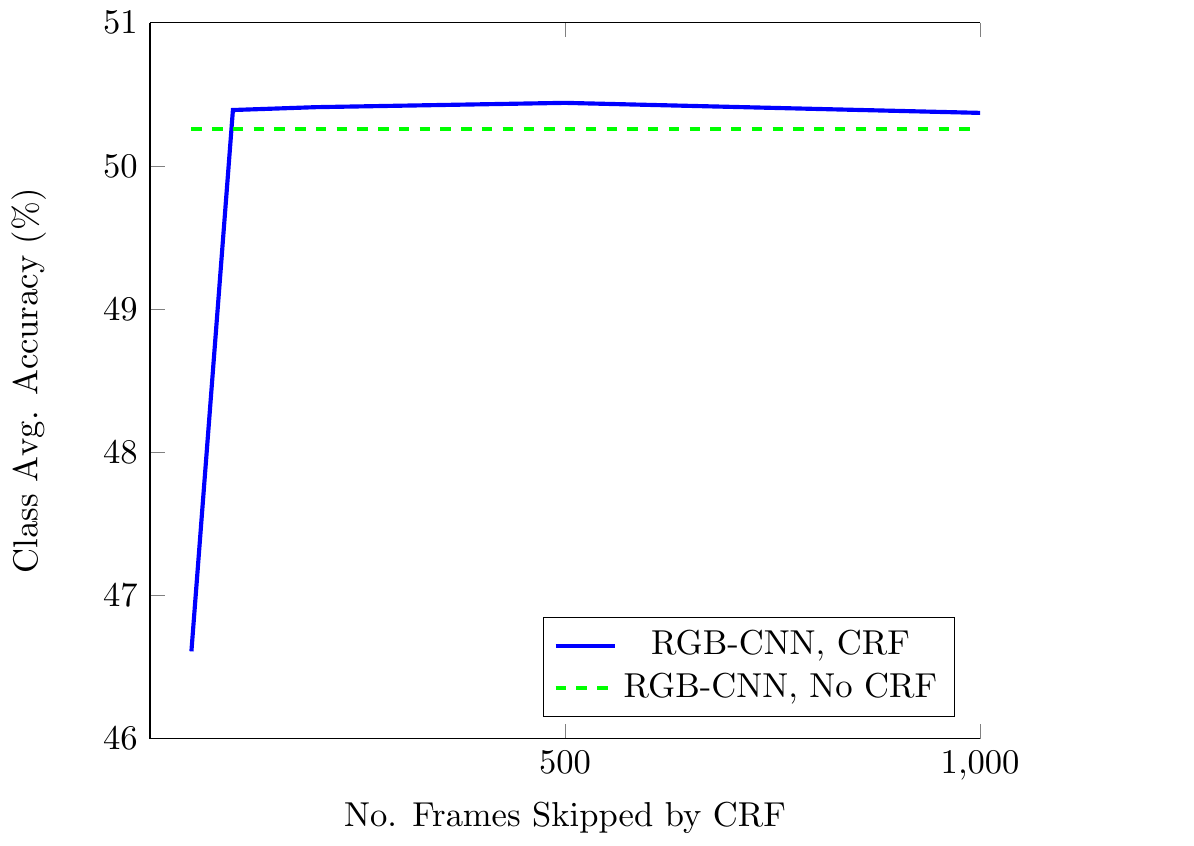}
\caption{The average class accuracy processing every 10\textsuperscript{th} frame with a CNN, with a variable number of frames between CRF updates.  If applied too frequently the CRF was detrimental to performance, and the performance improvement from the CRF was not significant for this CNN.}
\vspace{2mm}\hrule 
\label{fig:crf_frameskip_vs_accuracy}
\end{figure}

We also evaluated the effect of varying the number of frames between CRF updates (Figure~\ref{fig:crf_frameskip_vs_accuracy}).  We found that when applied too frequently, the CRF can `drown out' predictions of the CNN, resulting in a significant reduction in accuracy.  Performing an update every 500 frames results in a slight improvement, and so we use that as the default update rate in all subsequent experiments.

\subsection{Accuracy Evaluation}

We evaluate the accuracy of our SemanticFusion pipeline against the accuracy achieved by a single frame CNN segmentation. The results of this evaluation are summarised in Table~\ref{table:office_reconstruction}. We observe that in all cases semantically fusing additional viewpoints improved the accuracy of the segmentation over a single frame system. Performance improved from 43.6\% for a single frame to 48.3\% when projecting the predictions from the 3D SemanticFusion map. 

We also evaluate our system on the office dataset when using predictions from the state-of-the-art CNN developed by Eigen \textit{et al.}\footnote{We use the publicly available network weights and implementation from: \texttt{http://www.cs.nyu.edu/\textasciitilde deigen/dnl/.}} based on the VGG architecture.  To maintain consistency with the rest of the system, we perform only a single forward pass of the network to calculate the output probabilities.  The network requires ground truth normal information, and so to ensure the input pipeline is the same as in Eigen \textit{et al.}~\cite{Eigen:etal:ICCV2015}, we preprocess the sequence with the MATLAB script linked to in the project page to produce the ground truth normals.  With this setup we see an improvement of 2.9\% over the single frame implementation with SemanticFusion, from 57.1\% to 60.0\%.

The performance benefit of the CRF was less clear.  It provided a very small improvement of 0.5\% for the Eigen network, but a slight detriment to the RGBD-CNN of 0.2\%.

\begin{table*}
\begin{tabular}{l}
\textbf{Office Reconstruction: 13 Class Semantic Segmentation} \\ 
\end{tabular}
\centering
\begin{tabular}{|l|p{0.4cm}|p{0.4cm}|p{0.4cm}|p{0.4cm}|p{0.4cm}|p{0.4cm}|p{0.4cm}|p{0.4cm}|p{0.4cm}|p{0.4cm}|p{0.4cm}|p{0.4cm}}
\hline
Method  &  \cellcolor{booksColor}\rotatebox{90}{books}  & \cellcolor{ceilColor}\rotatebox{90}{\textcolor{white}{ceiling}} & \cellcolor{chairColor}\rotatebox{90}{\textcolor{white}{chair}}  & \cellcolor{floorColor}\rotatebox{90}{floor}  & \cellcolor{objsColor}\rotatebox{90}{objects} & \cellcolor{paintColor}\rotatebox{90}{\textcolor{white}{painting}}   & \cellcolor{tableColor}\rotatebox{90}{table}  & \cellcolor{wallColor}\rotatebox{90}{wall}   & \cellcolor{windColor}\rotatebox{90}{window} & \rotatebox{90}{class avg. } & \rotatebox{90}{pixel avg. } \\ \hline
\rule{0pt}{4ex}RGBD &  61.8 & 48.2 & 28.6 & 63.9 & 41.8 & 39.5  & 9.1 & 80.6 & 18.9 & 43.6 & 47.0  \\ \hline
RGBD-SF & \textbf{66.4 }& \textbf{78.7} & 36.8 & 63.4 & 41.9 & 26.2 & 12.1 & 84.2 & 25.3 & 48.3 & 54.7 \\ \hline
RGBD-SF-CRF & \textbf{66.4} & 78.1 & 37.2 & 64.2 & 40.8 & 27.5 & 10.6 & 85.1 & 22.7 & 48.1 & 54.8 \\ \hline
\rule{0pt}{4ex}Eigen~\cite{Eigen:etal:ICCV2015} & 57.8 & 54.3 & 57.8 & 72.8 & 49.4 & 77.5 & 24.1 & 81.6 & 38.9 & 57.1 & 62.5 \\ \hline
Eigen-SF &60.8 & 58.0 & 62.8 & 74.9 & \textbf{53.3} & \textbf{80.3}& \textbf{24.6} & 86.3 & 38.8 & 60.0 & 65.8 \\ \hline
Eigen-SF-CRF & 65.9 & 53.3 & \textbf{65.1} & \textbf{76.8}  & 53.1 & 79.6  & 22.0  & \textbf{87.7} & \textbf{41.4} & \textbf{60.5} & \textbf{67.0} \\ \hline

\end{tabular}
\vspace{0.5mm} \vspace{0.5mm}
\caption{\textbf{Reconstruction dataset results}:  \small SF denotes that the labels were produced by SemanticFusion, and the results where captured immediately if a frame with ground truth labelling was present. When no reconstruction is present for a pixel, we fall back to the predictions of the baseline single frame network. All accuracy evaluations were performed at $320\times240$ resolution.} 
\label{table:office_reconstruction}
\vspace{2mm}\hrule 
\end{table*}

\subsection{NYU Dataset}

We choose to validate our approach on the NYUv2 dataset~\cite{Silberman:etal:ECCV2012}, as it is one of the few datasets which provides all of the information required to evaluate semantic RGB-D reconstruction.  The SUN RGB-D \cite{Song:etal:CVPR2015}, although an order of magnitude larger than NYUv2 in terms of labelled images, does not provide the raw RGB-D videos and therefore is could not be used in our evaluation. 

\begin{figure}
\centering
\includegraphics[width=8.2cm]{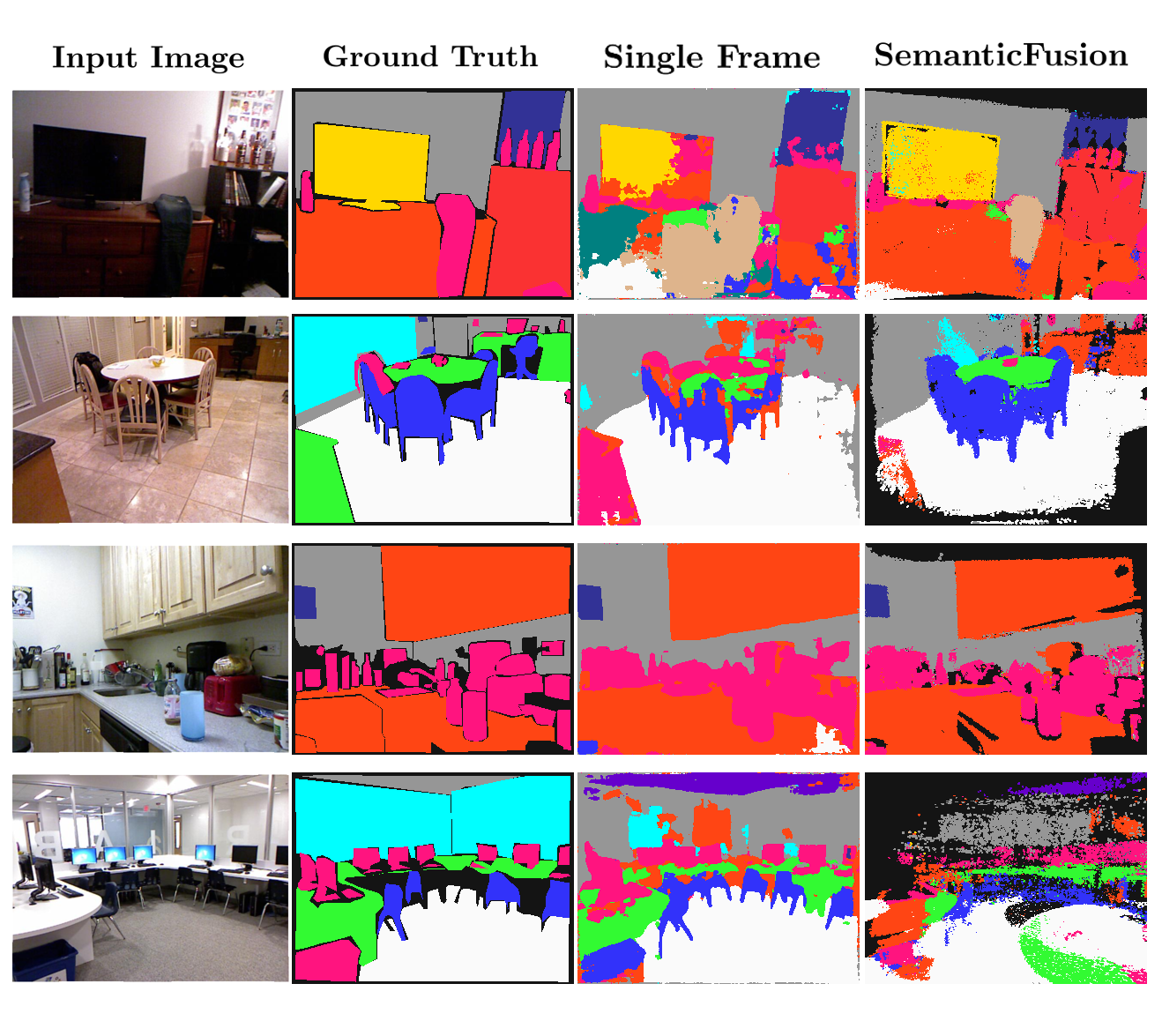}
\caption{\textbf{Qualitative NYUv2 test set results}: \small The results of SemanticFusion are using the RGBD-CNN with CRF after the completed trajectory, against the same networks single frame predictions.  For evaluation, the black regions of SemanticFusion denoting areas without a reconstruction, are replaced with the baseline CNN predictions. Here we show only the semantic reconstruction result for clarity.  The first two rows show instances where SemanticFusion has clearly improved the accuracy of the 2D annotations.  The third row shows an example of a very rotational trajectory, where there is little difference as a result of fusing predictions.  The final row shows an example where the trajectory was clearly not taken with reconstruction in mind, and the distant geometry leads to tracking and mapping problems even within our subset requiring 2Hz frame-rate.  Cases such as this provide an advantage to the accuracy of the single frame network.}
\vspace{2mm}\hrule 
\label{fig:nyu_qualitative}
\end{figure}

The NYUv2 dataset itself is still not ideally suited to the role.  Many of the 206 test set video sequences exhibit significant drops in frame-rate and thus prove unsuitable for tracking and reconstruction.  In our evaluations we excluded any sequence which experienced a frame-rate under 2Hz. The remaining 140 test sequences result in 360 labelled test images of the original 654 image test set in NYUv2.  The results of our evaluation are presented in Table~\ref{table:nyu_quantitative} and some qualitative results are shown in Figure~\ref{fig:nyu_qualitative}.  

\begin{table*}[!htbp]
\begin{tabular}{l}
\textbf{NYUv2 Test Set: 13 Class Semantic Segmentation} \\ 
\end{tabular}
\centering
\begin{tabular}{|l|p{0.4cm}|p{0.4cm}|p{0.4cm}|p{0.4cm}|p{0.4cm}|p{0.4cm}|p{0.4cm}|p{0.4cm}|p{0.4cm}|p{0.4cm}|p{0.4cm}|p{0.4cm}|p{0.4cm}|p{0.4cm}|p{0.4cm}|p{0.4cm}}
\hline
Method  &  \cellcolor{bedColor}\rotatebox{90}{\textcolor{white}{bed}}&  \cellcolor{booksColor}\rotatebox{90}{books}  & \cellcolor{ceilColor}\rotatebox{90}{\textcolor{white}{ceiling}} & \cellcolor{chairColor}\rotatebox{90}{\textcolor{white}{chair}}  & \cellcolor{floorColor}\rotatebox{90}{floor}  & \cellcolor{furnColor}\rotatebox{90}{furniture}& \cellcolor{objsColor}\rotatebox{90}{objects} & \cellcolor{paintColor}\rotatebox{90}{\textcolor{white}{painting}}   & \cellcolor{sofaColor}\rotatebox{90}{sofa}& \cellcolor{tableColor}\rotatebox{90}{table}  & \cellcolor{tvColor}\rotatebox{90}{tv}&\cellcolor{wallColor}\rotatebox{90}{wall}   & \cellcolor{windColor}\rotatebox{90}{window} & \rotatebox{90}{class avg. } & \rotatebox{90}{pixel avg. } \\ \hline
\rule{0pt}{4ex}RGBD & 62.5 & \textbf{60.5} & 35.0 & 51.7 & 92.1 & 54.5 & \textbf{61.3} & \textbf{72.1} & 34.7 & 26.1 & 32.4 & 86.5 & 53.5 & 55.6 & 62.0  \\ \hline
RGBD-SF & 61.7 & 58.5 & 43.4 & 58.4 & 92.6 & 63.7 & 59.1 & 66.4 & 47.3 & 34.0 & 33.9 & 86.0 & 60.5 & 58.9 & 67.5 \\ \hline
RGBD-SF-CRF & 62.0 & 58.4 & 43.3 & 59.5 & \textbf{92.7}& \textbf{64.4} & 58.3 & 65.8 & \textbf{48.7} & 34.3 & 34.3 & 86.3 & \textbf{62.3} & 59.2 & 67.9  \\ \hline
\rule{0pt}{4ex}Eigen~\cite{Eigen:etal:ICCV2015} &42.3 & 49.1 & 73.1 & 72.4 & 85.7 & 60.8 & 46.5 & 57.3 & 38.9 & 42.1 & 68.5 & 85.5 & 55.8 & 59.9 & 66.5 \\ \hline
Eigen-SF & 47.8 & 50.8 & 79.0 & 73.3 & 90.5 & 62.8 & 46.7 & 64.5 & 45.8 & \textbf{46.0} & 70.7 & 88.5 & 55.2 & 63.2 & 69.3 \\ \hline
Eigen-SF-CRF  &48.3 & 51.5 & 79.0 & \textbf{74.7} & 90.8 & 63.5 & 46.9 & 63.6 & 46.5 & 45.9 & \textbf{71.5} & \textbf{89.4} & 55.6 & \textbf{63.6} & \textbf{69.9} \\ \hline
\rule{0pt}{4ex}Hermans \textit{et al.}~\cite{Hermans:etal:ICRA2014} & \textbf{68.4} & 45.4  & \textbf{83.4} & 41.9 & 91.5 & 37.1 & 8.6 & 35.8 & 28.5 & 27.7 & 38.4& 71.8 & 46.1 &  48.0 & 54.3 \\ \hline
\end{tabular}
\vspace{0.5mm} \vspace{0.5mm}
\caption{\textbf{NYUv2 test set results}: \small SF denotes that the labels were produced by SemanticFusion, and the results were captured immediately if a keyframe was present. When no reconstruction is present for a pixel, we fall back to the predictions of the baseline single frame network. Note that we calculated the accuracies of \cite{Eigen:etal:ICCV2015} using their publicly available implementation. Our results are not directly comparable with Hermans \textit{et al.}~\cite{Hermans:etal:ICRA2014} as we only evaluate on a subset of the test set, and their annotations are not available.  However, we include their results for reference. Following previous work~\cite{Hermans:etal:ICRA2014} we exclude pixels without a corresponding depth measurement. All accuracy evaluations were performed at $320\times240$ resolution.} 
\label{table:nyu_quantitative}
\vspace{2mm}\hrule 
\end{table*}

Overall, fusing semantic predictions resulted in a notable improvement over single frame predictions. However, the total relative gains of 2.3\%  for the RGBD-CNN was approximately half of the 4.7\% improvement witnessed in the office reconstruction dataset. We believe this is largely a result of the style of capturing NYUv2 datasets. The primarily rotational scanning pattern often used in test trajectories does not provide as many useful different viewpoints from which to fuse independent predictions.  Despite this, there is still a significant accuracy improvement over the single frame predictions. 

We also improved upon the state-of-the-art Eigen \textit{et al.}~\cite{Eigen:etal:ICCV2015} CNN, with the class average accuracy going from 59.9\% to 63.2\% (\verb!+!3.3\%). This result clearly shows, even on this challenging dataset, the capacity of SemanticFusion to not only provide a useful semantically annotated 3D map, but also to improve the predictions of state-of-the-art 2D semantic segmentation systems. 

The improvement as a result of the CRF was not particularly significant, but positive for both CNNs.  Eigen's CNN saw \verb!+!0.4\% improvement, and the RGBD-CNN saw \verb!+!0.3\%.  This could possibly be improved with proper tuning of edge potential weights and unit standard deviations, and the potential exists to explore many other kinds of map-based semantic regularisation schemes. We leave these explorations to future work.

\subsection{Run-time Performance}
\label{sec:runtime_analysis}

We benchmark the performance of our system on a random sample of 30 sequences from the NYUv2 test set.  All tests were performed on an Intel Core i7-5820K 3.30GHz CPU and an NVIDIA Titan Black GPU.  Our SLAM system requires 29.3ms on average to process each frame and update the map.  For every frame we also update our stored surfel probability table to account for any surfels removed by the SLAM system. This process requires an additional 1.0ms. As discussed above, the other components in our system do not need to be applied for every frame. A forward pass of our CNN requires 51.2ms and our Bayesian update scheme requires a further 41.1ms.  Our standard scheme performs this every 10 frames, resulting in an average frame-rate of 25.3Hz.

Our experimental CRF implementation was developed only for the CPU in C\verb!++!, but the message passing algorithm adopted could lend itself to an optimised GPU implementation. The overhead of copying data from the GPU and performing inference on a single threaded CPU implementation is significant.  Therefore on average, it takes 20.3s to perform 10 CRF iterations.  In the evaluation above, we perform a CRF update once every 500 frames, but for online use it can be disabled entirely or applied once at the conclusion of a sequence.

\section{CONCLUSIONS}

Our results confirm the strong expectation that using a SLAM system to provide pixel-wise correspondences between frames allows the fusion of per-frame 2D segmentations into a coherent 3D semantic map. It is the first time that this has been demonstrated with a real-time, loop-closure capable approach suitable for interactive room scanning. Not only that, the incorporation of such a map led to a significant improvement in the corresponding 2D segmentation accuracy.

We exploited the flexibility of CNNs to improve the accuracy of a pretrained RGB network by incoporating an additional depth channel.  In this work we opted for the simplest feasible solution to allow this new modality.  Some recent work has explored other ways to incorporate depth information~\cite{Hoffman:etal:ICRA2016}, but such an approach requires a duplication of the lower network parameters and was infeasible in our system due to GPU memory limitations. However, future research could also incorporate recent breakthroughs in CNN compression~\cite{Iandola:etal:ARXIV2016}, which would not only enable improvements to the incorporation of other modalities, but also offer exciting new directions to enable real-time semantic segmentation on low memory and power mobile devices.

We believe that this is just the start of how knowledge from SLAM and machine-learned labelling can be brought
together to enable truly powerful semantic and object-aware mapping. Our own reconstruction-focused dataset shows a much larger improvement in labelling accuracy via fusion than the NYU dataset with less varied trajectories, this underlines the importance of viewpoint variation.  It also hints at the improvements that could be achieved with significantly longer trajectories, such as those of an autonomous robot in the field making direct use of the semantically annotated 3D map.

Going further, it is readily apparent, as demonstrated in a so far relatively simple manner in systems like SLAM\verb!++!~\cite{Salas-Moreno:etal:CVPR2013} that not just should reconstruction be used to provide correspondence to help labelling, but that labelling/recognition can make reconstruction and SLAM much more accurate and efficient. A loop-closure capable surfel map as in ElasticFusion is highly suitable for applying operations such as class-specific smoothing (as in the extreme case of planar region recognition and fitting~\cite{Salas-Moreno:etal:ISMAR2014}), and this will be an interesting direction. More powerful still will be to interface with explicit object instance recognition and to replace elements of the surfel model directly with 3D object models once confidence reaches a suitable threshold.

\bibliographystyle{IEEEtranS}
\bibliography{robotvision}

\begin{thebibliography}{10}
\providecommand{\url}[1]{#1}
\csname url@rmstyle\endcsname
\providecommand{\newblock}{\relax}
\providecommand{\bibinfo}[2]{#2}
\providecommand\BIBentrySTDinterwordspacing{\spaceskip=0pt\relax}
\providecommand\BIBentryALTinterwordstretchfactor{4}
\providecommand\BIBentryALTinterwordspacing{\spaceskip=\fontdimen2\font plus
\BIBentryALTinterwordstretchfactor\fontdimen3\font minus
  \fontdimen4\font\relax}
\providecommand\BIBforeignlanguage[2]{{%
\expandafter\ifx\csname l@#1\endcsname\relax
\typeout{** WARNING: IEEEtran.bst: No hyphenation pattern has been}%
\typeout{** loaded for the language `#1'. Using the pattern for}%
\typeout{** the default language instead.}%
\else
\language=\csname l@#1\endcsname
\fi
#2}}

\bibitem{Couprie:etal:ICLR2013}
C.~Couprie, C.~Farabet, L.~Najman, and Y.~{LeCun}, ``Indoor semantic
  segmentation using depth information,'' in \emph{{Proceedings of the
  International Conference on Learning Representations ({ICLR})}}, 2013.

\bibitem{Eigen:etal:ICCV2015}
D.~Eigen and R.~Fergus, ``{Predicting Depth, Surface Normals and Semantic
  Labels with a Common Multi-Scale Convolutional Architecture},'' in
  \emph{{Proceedings of the International Conference on Computer Vision
  ({ICCV})}}, 2015.

\bibitem{Everingham:etal:IJCV2010}
M.~Everingham, L.~Van~Gool, C.~K. Williams, J.~Winn, and A.~Zisserman, ``The
  pascal visual object classes ({VOC}) challenge,'' \emph{{International
  Journal of Computer Vision ({IJCV})}}, no.~2, pp. 303--338, 2010.

\bibitem{Gupta:etal:ECCV2014}
S.~Gupta, R.~Girshick, P.~Arbelaez, and J.~Malik, ``{Learning Rich Features
  from {RGB-D} Images for Object Detection and Segmentation},'' in
  \emph{{Proceedings of the European Conference on Computer Vision ({ECCV})}},
  2014.

\bibitem{Gupta:etal:CVPR2015B}
S.~Gupta, P.~A. Arbel{\'{a}}ez, R.~B. Girshick, and J.~Malik, ``Aligning {3D}
  models to {RGB-D} images of cluttered scenes,'' in \emph{{Proceedings of the
  {IEEE} Conference on Computer Vision and Pattern Recognition ({CVPR})}},
  2015.

\bibitem{Gupta:etal:CVPR2015}
S.~Gupta, P.~A. Arbel{\'{a}}ez, R.~B. Girshick, and J.~Malika, ``{Aligning 3D
  Models to RGB-D Images of Cluttered Scenes},'' in \emph{{Proceedings of the
  {IEEE} Conference on Computer Vision and Pattern Recognition ({CVPR})}},
  2015.

\bibitem{Handa:etal:ARXIV2015}
A.~Handa, V.~P{\u a}tr{\u a}ucean, V.~Badrinarayanan, S.~Stent, and R.~Cipolla,
  ``Scene{N}et: Understanding {R}eal {W}orld {I}ndoor {S}cenes {W}ith
  {S}ynthetic {D}ata,'' \emph{{arXiv preprint arXiv}:1511.07041}, 2015.

\bibitem{Hermans:etal:ICRA2014}
A.~Hermans, G.~Floros, and B.~Leibe, ``Dense 3d semantic mapping of indoor
  scenes from rgb-d images,'' in \emph{{Proceedings of the {IEEE} International
  Conference on Robotics and Automation ({ICRA})}}, 2014.

\bibitem{Hoffman:etal:ICRA2016}
J.~Hoffman, S.~Gupta, J.~Leong, G.~S., and T.~Darrell, ``{Cross-Modal
  Adaptation for RGB-D Detection},'' in \emph{{Proceedings of the {IEEE}
  International Conference on Robotics and Automation ({ICRA})}}, 2016.

\bibitem{Iandola:etal:ARXIV2016}
F.~N. Iandola, M.~W. Moskewicz, K.~Ashraf, S.~Han, W.~J. Dally, and K.~Keutzer,
  ``Squeezenet: Alexnet-level accuracy with 50x fewer parameters and
  {\textless}1mb model size,'' \emph{CoRR}, 2016.

\bibitem{Jia:etal:ARXIV2014}
Y.~Jia, E.~Shelhamer, J.~Donahue, S.~Karayev, J.~Long, R.~Girshick,
  S.~Guadarrama, and T.~Darrell, ``Caffe: Convolutional architecture for fast
  feature embedding,'' \emph{arXiv preprint arXiv:1408.5093}, 2014.

\bibitem{Koppula:etal:NIPS2011}
H.~S. Koppula, A.~Anand, T.~Joachims, and A.~Saxena, ``{Semantic Labeling of 3D
  Point Clouds for Indoor Scenes},'' in \emph{{Neural Information Processing
  Systems ({NIPS})}}, 2011.

\bibitem{Krahenbuhl:Koltun:NIPS2011}
P.~Kr\"{a}henb\"{u}hl and V.~Koltun, ``{Efficient Inference in Fully Connected
  CRFs with Gaussian Edge Potentials},'' in \emph{{Neural Information
  Processing Systems ({NIPS})}}, 2011.

\bibitem{Krizhevsky:etal:NIPS2012}
A.~Krizhevsky, I.~Sutskever, and G.~Hinton, ``{ImageNet} classification with
  deep convolutional neural networks,'' in \emph{{Neural Information Processing
  Systems ({NIPS})}}, 2012.

\bibitem{Lin:etal:ECCV2014}
T.-Y. Lin, M.~Maire, S.~Belongie, J.~Hays, P.~Perona, D.~Ramanan,
  P.~Doll{\'a}r, and C.~L. Zitnick, ``Microsoft {COCO}: Common objects in
  context,'' in \emph{{Proceedings of the European Conference on Computer
  Vision ({ECCV})}}, 2014, pp. 740--755.

\bibitem{Long:etal:CVPR2015}
J.~Long, E.~Shelhamer, and T.~Darrell, ``Fully convolutional networks for
  semantic segmentation,'' in \emph{{Proceedings of the {IEEE} Conference on
  Computer Vision and Pattern Recognition ({CVPR})}}, 2015.

\bibitem{Noh:etal:ARXIV2015}
H.~Noh, S.~Hong, and B.~Han, ``Learning deconvolution network for semantic
  segmentation,'' \emph{arXiv preprint arXiv:1505.04366}, 2015.

\bibitem{Salas-Moreno:etal:ISMAR2014}
R.~F. Salas-Moreno, B.~Glocker, P.~H.~J. Kelly, and A.~J. Davison, ``{Dense
  Planar SLAM},'' in \emph{{Proceedings of the International Symposium on Mixed
  and Augmented Reality ({ISMAR})}}, 2014.

\bibitem{Salas-Moreno:etal:CVPR2013}
\BIBentryALTinterwordspacing
R.~F. Salas-Moreno, R.~A. Newcombe, H.~Strasdat, P.~H.~J. Kelly, and A.~J.
  Davison, ``{{SLAM++}: Simultaneous Localisation and Mapping at the Level of
  Objects},'' in \emph{{Proceedings of the {IEEE} Conference on Computer Vision
  and Pattern Recognition ({CVPR})}}, 2013. [Online]. Available:
  \url{http://dx.doi.org/10.1109/CVPR.2013.178}
\BIBentrySTDinterwordspacing

\bibitem{Silberman:etal:ECCV2012}
N.~Silberman, D.~Hoiem, P.~Kohli, and R.~Fergus, ``Indoor segmentation and
  support inference from {RGBD} images,'' in \emph{{Proceedings of the European
  Conference on Computer Vision ({ECCV})}}, 2012.

\bibitem{Simonyan:Zisserman:ICLR2015}
K.~Simonyan and A.~Zisserman, ``{Very Deep Convolutional Networks for
  Large-Scale Image Recognition},'' in \emph{{Proceedings of the International
  Conference on Learning Representations ({ICLR})}}, 2015.

\bibitem{Song:etal:CVPR2015}
S.~Song, S.~P. Lichtenberg, and J.~Xiao, ``{SUN} {RGB-D}: A {RGB-D} scene
  understanding benchmark suite,'' in \emph{{Proceedings of the {IEEE}
  Conference on Computer Vision and Pattern Recognition ({CVPR})}}, 2015, pp.
  567--576.

\bibitem{Stuckler:etal:JRTIP2015}
J.~St\"uckler, B.~Waldvogel, H.~Schulz, and S.~Behnke, ``Multi-resolution
  surfel maps for efficient dense 3d modeling and tracking,'' \emph{{Journal of
  Real-Time Image Processing {JRTIP}}}, vol.~10, no.~4, pp. 599--609, 2015.

\bibitem{Valentin:etal:CVPR2014}
J.~Valentin, S.~Sengupta, J.~Warrell, A.~Shahrokni, and P.~Torr, ``{Mesh Based
  Semantic Modelling for Indoor and Outdoor Scenes},'' in \emph{{Proceedings of
  the {IEEE} Conference on Computer Vision and Pattern Recognition ({CVPR})}},
  2013.

\bibitem{Whelan:etal:RSS2015}
T.~Whelan, S.~Leutenegger, R.~F. Salas-Moreno, B.~Glocker, and A.~J. Davison,
  ``{ElasticFusion}: Dense {SLAM} without a pose graph,'' in \emph{{Proceedings
  of Robotics: Science and Systems ({RSS})}}, 2015.

\end{thebibliography}

\end{document}